\documentclass[manuscript,screen]{acmart}
\usepackage{graphicx}%
\usepackage{multirow}%
\usepackage{amsmath,amsfonts}%
\usepackage{amsthm}%
\usepackage{mathrsfs}%
\usepackage[title]{appendix}%
\usepackage{xcolor}%
\usepackage{textcomp}%
\usepackage{manyfoot}%
\usepackage{booktabs}%
\usepackage{algorithm}%
\usepackage{algorithmicx}%
\usepackage{algpseudocode}%
\usepackage{listings}%
\usepackage{subfig}
\usepackage{changepage}
\usepackage{wrapfig}
\AtBeginDocument{%
  }

\setcopyright{acmlicensed}
\copyrightyear{2025}
\acmYear{2025}
\acmDOI{XXXXXXX.XXXXXXX}

\begin{document}

\title{Generative AI Meets Future Cities: Towards an Era of Autonomous Urban Intelligence}

\author{Dongjie Wang}
\orcid{https://orcid.org/0000-0003-3948-0059}
\affiliation{%
  \department{Electrical Engineering and Computer Science}
  \institution{University of Kansas}
  \city{Lawrence}
  \country{USA}
}
\email{wangdongjie@ku.edu}

\author{Chang-Tien Lu}
\orcid{https://orcid.org/0000-0003-3675-0199}
\affiliation{%
  \department{Department of Computer Science}
  \institution{Virginia Tech}
  \city{Falls Church}
  \country{USA}}
\email{ctlu@vt.edu}

\author{Xinyue Ye}
\orcid{https://orcid.org/0000-0001-8838-9476}
\affiliation{%
  \department{Department of Geography \& the Environment}
  \institution{University of Alabama}
  \city{Tuscaloosa}
  \country{USA}
}
\email{xye10@ua.edu}

\author{Tan Yigitcanlar}
\orcid{https://orcid.org/0000-0001-7262-7118}
\affiliation{
  \institution{Queensland University of Technology}
  \city{Brisbane}
  \country{Australia}
}
\email{tan.yigitcanlar@qut.edu.au}

\author{Yanjie Fu}
\orcid{https://orcid.org/0000-0002-1767-8024}
\authornotemark[2]
\affiliation{%
  \department{School of Computing and Augmented Intelligence}
  \institution{Arizona State University}
  \city{Tempe}
  \country{USA}
}
\email{yanjie.fu@asu.edu}

\renewcommand{\shortauthors}{Wang and Lu et al.}

\begin{abstract}
The two fields of urban planning and artificial intelligence (AI) arose and developed separately over time. There is now, however, cross-pollination and growing interest in their intersection. This paper highlights the importance of urban planning from the perspectives of quality of life, economic development, disaster risk reduction, and environmental sustainability. The paper reviews fundamental urban planning concepts and relate them to key open problems in AI technologies—particularly machine learning—including adversarial learning, generative neural networks, deep encoder-decoder networks, conversational AI, and geospatial and temporal machine learning—thereby examining how AI can contribute to contemporary urban planning. A central issue this paper addresses is automated land-use configuration, formulated as the generation of land uses and building configurations for a target area based on geospatial data, mobility patterns, social media insights, environmental factors, and economic activities. The paper also outlines key implications of AI for urban planning and propose critical research directions toward automated urban planning.
\end{abstract}




\maketitle

\section{When Urban Planning Meets Artificial Intelligence}

Urban planning is the process of designing and managing the physical and social development of cities, towns, and other urban areas. It involves creating and implementing plans and policies to guide the use of land, infrastructure, and resources in a way that meets the needs of the community. This encompasses a wide range of activities, such as land use planning, transportation planning, environmental planning, economic development planning, and community development~\cite{goetz2020whiteness}.

Urban planning plays an indispensable role in shaping the future of our cities, and, consequently, the quality of life for millions of people. As cities continue to grow and evolve, urban planning becomes increasingly crucial for ensuring sustainable development, efficient use of resources, and fostering economic growth. Thoughtful urban planning creates environmentally responsible and socially inclusive communities, which not only enhance the quality of life for residents but also bolster the resilience of cities against natural disasters and other unforeseen challenges. By integrating cutting-edge technologies, such as deep generative learning, into the urban planning process, we can better anticipate and respond to the complex, interconnected challenges of urbanization, while simultaneously leveraging opportunities for innovation and growth. Ultimately, this paves the way for more intelligent, adaptive, and resilient urban spaces that are better equipped to face the demands of a rapidly changing world.

The traditional urban planning process typically involves several stages, which may vary depending on the specific context and requirements of the project. Initially, the process entails assessment and analysis of the existing conditions of the urban area, as well as the social, economic, and environmental factors influencing its development. This is followed by goal setting, where desired outcomes and criteria for measuring progress are established. Potential strategies and interventions that can achieve the project's goals and objectives are identified during the strategy development stage. Once strategies are determined, a detailed plan outlining the specific actions, policies, and programs necessary for implementation is created. The implementation stage involves executing the plan by carrying out the outlined steps, policies, and programs, while collaborating with various stakeholders, securing funding, and managing resources. Finally, the process concludes with monitoring and evaluation to determine the plan's effectiveness and identify areas for improvement. Overall, the traditional urban planning process emphasizes a comprehensive and iterative approach, fostering collaboration with stakeholders and a deep understanding of the social, economic, and environmental factors influencing urban development.

\begin{wrapfigure}{r}{0.6\textwidth}
    \centering
    \includegraphics[width=0.8\linewidth]{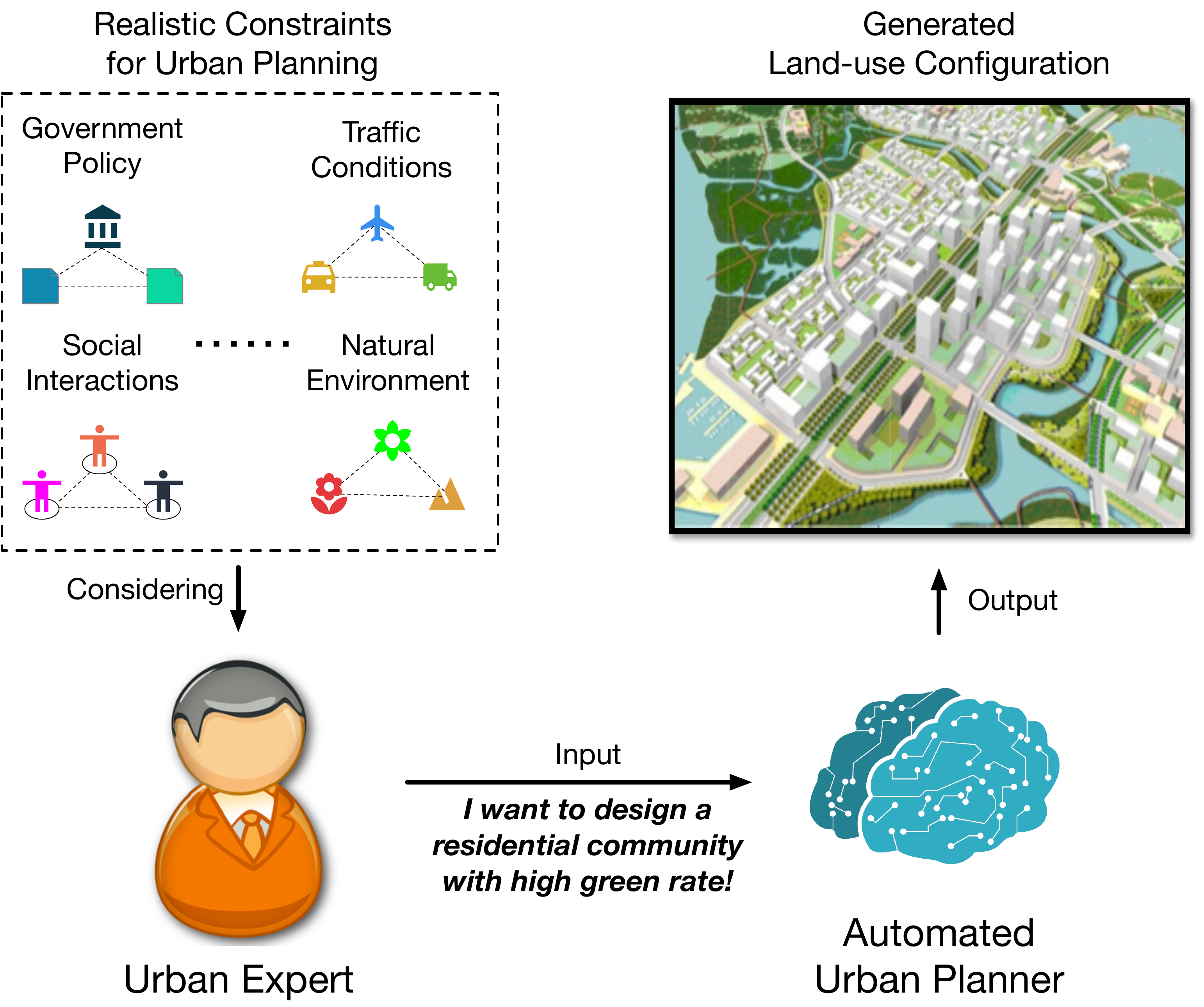}
    \caption{An AI-based urban planner generates optimal land-use configurations by taking into account  realistic planning constraints and the input of human urban experts.}
    \label{ai_generation}
\end{wrapfigure}

In recent years, advances in AI have opened up new opportunities for urban planners~\cite{ye2025artificial}. \textbf{Figure \ref{ai_generation}} shows that, by leveraging AI algorithms and data analysis techniques, planners can gain deeper insights into  complex systems (e.g., land uses, and transportation) that make up contemporary cities. This can help them to make more informed decisions about issues in urban planning. Moreover, AI can help to address some of the limitations of traditional urban planning methods. For example, AI can allow planners to process large amounts of geospatial and social data quickly and efficiently in near real time. Additionally, AI can help to identify patterns and trends that might be difficult or impossible to detect using traditional methods.

Despite the potential benefits of AI in urban planning, there are challenges and concerns to consider.
Owing to the high complexity of urban systems, planning tasks are predominantly undertaken by professional planners. However, human planners often require extended periods to complete these tasks. The recent advancements in Artificial Intelligence  Generative Content (AIGC) (e.g., ChatGPT, Stable Diffusion) offer significant potential for training machines to conceptualize and generate the land-use configuration plans of urban communities. This observation prompts us to reconsider urban planning from the perspective of AI: What roles does machine learning assume in urban planning? Can machines acquire and demonstrate human-like capabilities to autonomously and rapidly compute land-use configurations? 
By adopting this approach, could machines  serve as planning assistants, enabling human planners to adjust machine-generated plans to accommodate specific requirements?

\textit{Our Contributions:} 
In the present paper, we introduce a deep generative AI-driven urban planning framework that generates optimal land-use configurations, taking into account both planning constraints and expert input. This approach seeks to enhance our comprehension of the generative planning process and address existing limitations in urban planning methodologies, as illustrated in Figure~\ref{ai_generation}.
This would take the urban planning field a step closer to a form of automated and generative planning that involves thinking in the sense of image generation. 
Despite the success of generative AI in text and image generation, cities are a complex ecosystem that involves geospatial, social, economic, and human dimensions. Integrating both urban planning and AI has been an enormous challenge in terms of neural architecture, training algorithms, complex and imperfect data, and computing. The present work reviews and synthesizes key contributions that have been made to this end:

\begin{itemize}
    \item We describe  deep generative urban planning, an aspiration towards automated urban planning in Section 2. We discuss the computing thinking behind the analogy of urban planning and generative AI, present key terminologies used in converting urban planning into a deep generative learning task, and finally, present a generic representation-generation framework for urban planning. 
    \item We review existing approaches to learn deep generative models for urban planning in Section 3. We cover three categories of generative planning methods: automated urban planning via generative adversarial learning, automated urban planning via variational autoencoder, and automated urban planning via transformer. Besides, we discuss knowledge guidance and domain awareness in generative planning, such as spatial hierarchy structure knowledge in urban planning, overcoming sparsity of urban planning data, etc.
    \item We envision the future of deep generative urban planning research. We discuss the future automated urban planner from a ChatGPT point of view. We present our thinking on integrating generative AI, conversational AI, human feedback in the loop or human-machine collaborative planning  as one, and discuss several other problems such as better representation of an urban planning configuration and fairness in urban planning. 
\end{itemize}

\section{Deep Generative Urban Planning: Analogy, Terminologies, and Algorithmic Framework}

\begin{figure}[!thbp]
    \centering
    \includegraphics[width=0.75\linewidth]{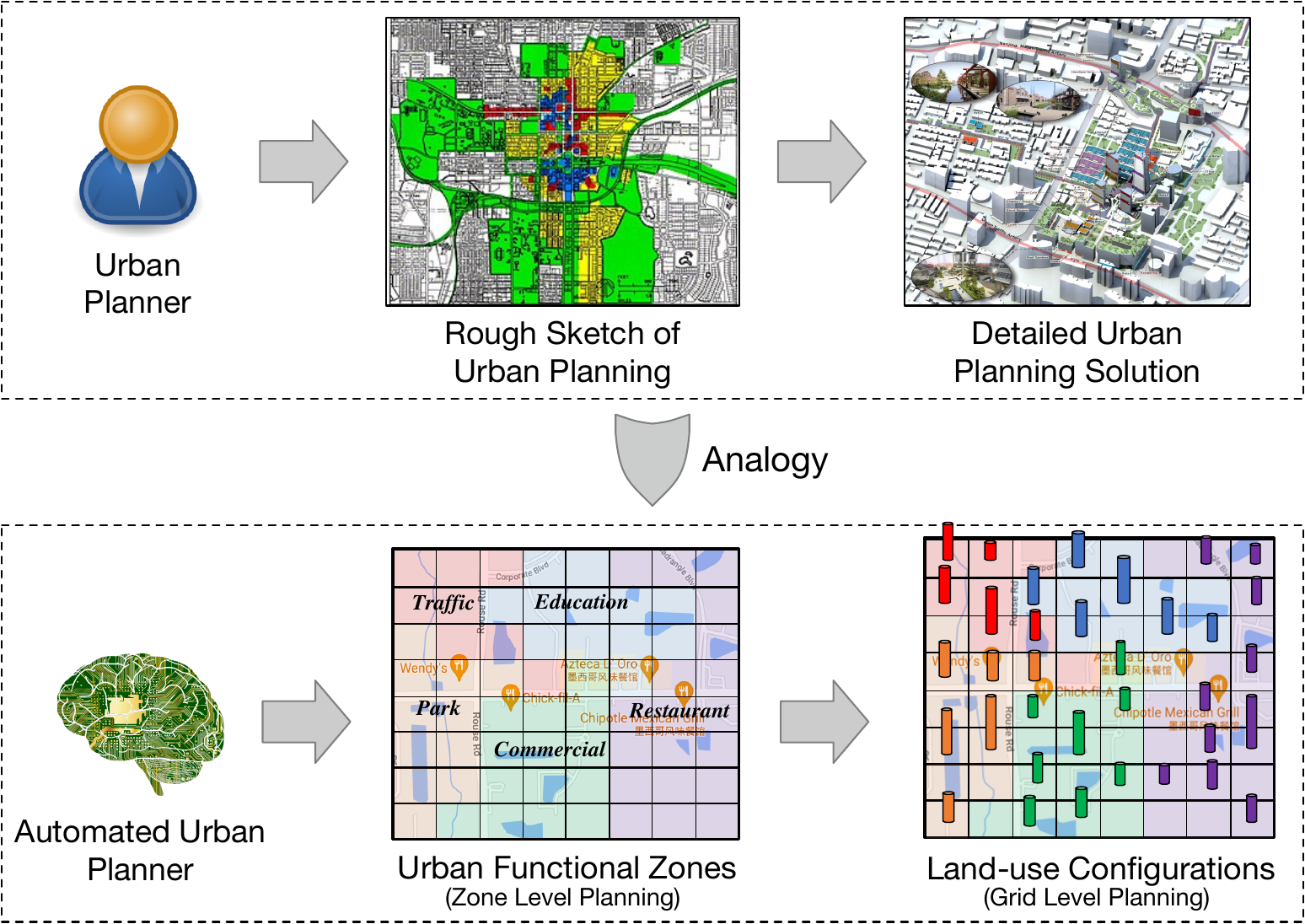}
      \vspace{-0.2cm}
    \caption{An analogy between urban planning and deep generative AI.}
    \label{generative}
    \vspace{-0.5cm}
\end{figure}

\subsection{A Generative AI Perspective: Automated Urban Planning as Deep Generative Learning}
Urban planning is a multidisciplinary field that involves designing and managing urban settlements. One interesting way to approach urban planning is to regard a community configuration as an image and regard planning a community configuration as generating an image. 
By thinking of urban planning as image generation (\textbf{Figure \ref{generative}}),  it allows us to formulate urban planning as a deep generative learning task. In other words, given a region or a community, a deep generative urban planning model can generate the optimal or near-optimal community geospatial configuration. 
Toward building such a deep generative urban planning framework, it is essential to address the three key questions:  1) Quantification: How can we quantify a land-use configuration plan? 2) Generation: How can we develop a machine learning framework that can learn the good and the bad of existing urban communities in terms of land-use configuration policies? 3) Evaluation: How can we evaluate the quality of generated land-use configurations? Before introducing technical details, we next present some key terminologies in deep generative urban planning. 

\subsection{Key Terminologies in Deep Generative Urban Planning}


\subsubsection{Target Area}
In urban planning, a target area refers to a specific geographic location that is the focus of planning efforts. This can be a community, a neighborhood, a district, an area, a city, or even a region, depending on the scale and scope of the planning activity. 
To make the target area quantifiable, we refer to a target area as a geographical region with an empty and square shape, typically denoted by a square of one-kilometer side length. 
In reality, target areas are typically identified based on a range of factors, including demographic characteristics, economic conditions, environmental features, transportation infrastructure, and social needs. 
Once a target area is identified, planners can use a variety of tools and techniques to analyze existing conditions and develop plans to improve the area. 

\begin{figure}[H]
\centering
\begin{adjustwidth}{-1cm}{-1cm}
\centering
\subfloat[\centering]{\includegraphics[width=4.3cm]{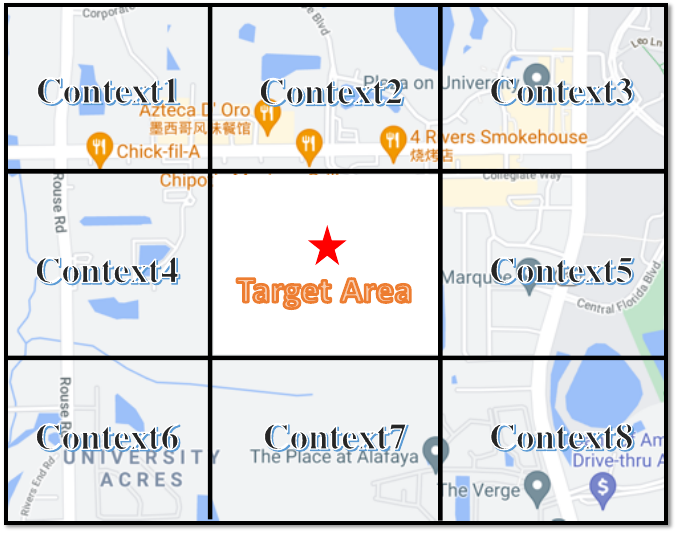}}
\hspace{10mm}
\subfloat[\centering]{\includegraphics[width=4.3cm]{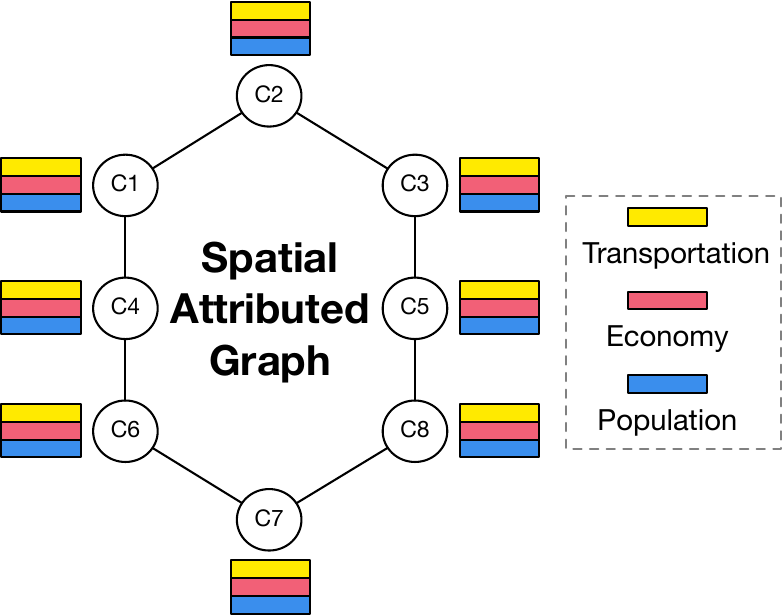}}
\end{adjustwidth}
\caption{Illustration of the target area and geospatial contexts. (\textbf{a}) Geospatial contexts encircle the target area from different directions.   (\textbf{b}) The spatial attributed graph contains all features of geospatial contexts.\label{target_geo}}
\end{figure}

\subsubsection{Geospatial Contexts}
Geospatial contexts represent the neighboring environments of a target area, with each context maintaining a shape akin to the target area.
\textbf{Figure \ref{target_geo} (a)} shows these contexts encircle the target area from different directions. In order to harness the information contained in geospatial contexts, as described in \cite{gan_planner}, we represent them as a spatial attributed graph, as illustrated in \textbf{Figure \ref{target_geo} (b)}. In this graph, a node denotes a geographical region, while an edge represents the spatial connectivity between any two regions. Additionally, each node is associated with the socioeconomic features of the corresponding region.

\subsubsection{Urban Functional Zone}
Urban functional zones, which provide a planning foundation for grid-level land-use configuration, are examined. Urban functional zones are extracted using geographical data and human mobility, as proposed in previous studies~\cite{yuan2014discovering}. To extract these zones, the geographical area is first divided into $N \times N$ grids, which are treated as "words," while human trajectories are considered "sentences." The collection of all trajectories within the geographical area is treated as a "document." A topic model is then employed to determine the urban function label of each grid. The resulting data structure of these zones is illustrated in \textbf{Figure ~\ref{urban_data} (a)}, where multiple grids are affiliated with a single urban function label in a matrix denoted by $\mathbf{U}\in \mathbb{R}^{N \times N}$.

\subsubsection{Land-use Configuration}
\begin{figure}[H]
\centering
\begin{adjustwidth}{-1cm}{-1cm}
\centering
\subfloat[\centering]{\includegraphics[width=4.3cm]{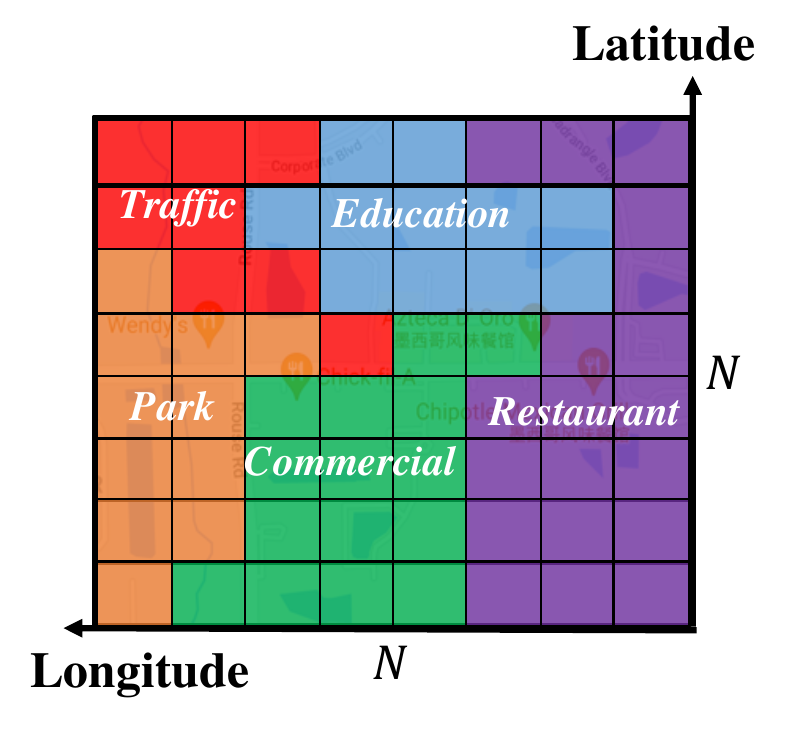}}
\hspace{10mm}
\subfloat[\centering]{\includegraphics[width=4.3cm]{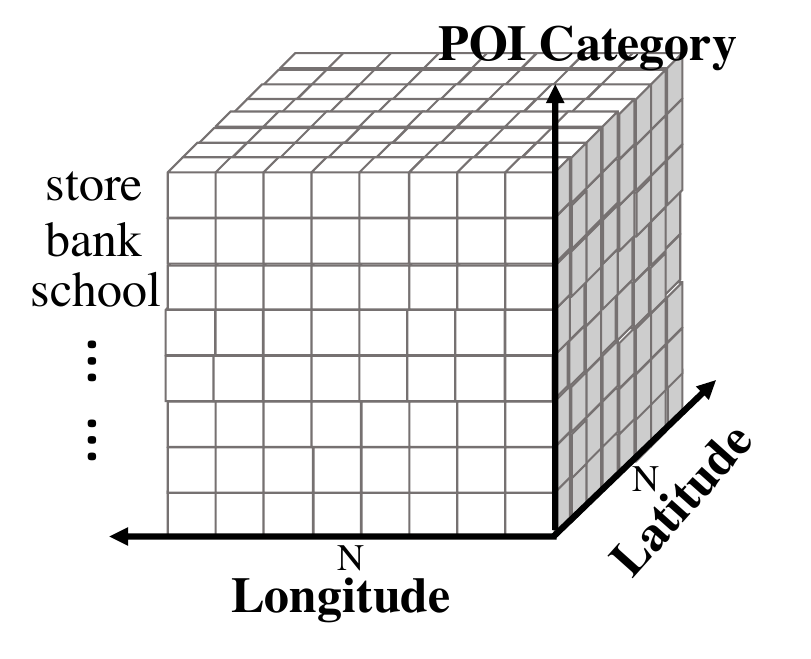}}
\end{adjustwidth}
\caption{Urban functional zone and land-use configuration. (\textbf{a}) Zone-level planning is a 2-D matrix, which provides high-level guidance for grid-level planning.   (\textbf{b}) Grid-level planning is represented by a 3-D tensor where we reserve the 3rd dimension for POI as each grid may contain multiple POI categories.\label{urban_data}}
\end{figure}

For land-use configuration, the quantitative definition introduced by ~\cite{gan_planner} is adopted. The geographical area is first partitioned into $N\times N$ grids, and the number of Points of Interest (POIs) within each grid is counted for each POI category. These counts of each category are then stacked to generate the final configuration. \textbf{Figure ~\ref{urban_data} (b)} shows the resulting data structure of this configuration, which is a tensor consisting of longitude, latitude, and POI category dimensions. This tensor is denoted by $\mathbf{\widehat{X}} \in \mathbb{R}^{N \times N \times C}$, where $C$ is the number of POI categories.

\subsubsection{Human Instruction}
Human instruction plays a crucial role in guiding the generation process of our planning framework. To enable our model to comprehend such instruction, we have devised a method to quantify its semantic meaning across different levels. For example, the range of green rate, which denotes the coverage of green plants within a geographical area, falls within the interval $[0\sim 1]$. We partition this interval into several green rate levels, with the label of each level reflecting the corresponding human instruction.

\subsection{A Generic Framework of Deep Generative Urban Planning}

Deep generative urban planning aims to generate the optimal  land-use configuration of a target area.
An automated urban planner involves two essential steps: representation and generation. 
The first step is to learn the representation of the geospatial, mobility, economic, social contexts and human instructions of a target area as generation conditions. 
The second step is to generate an optimal land use configuration solution for the target area given the representation of generation conditions. 


\subsubsection{Representation: Perceiving Geospatial Forms, Human Mobility,  Social Interactions, and Human Planner Instructions}
To generate effective urban land-use configuration, we need to teach AI to perceive and understand geospatial forms, human mobility, social interactions, and human planner instructions.

\noindent\underline{\emph{What to Represent.}} First, geospatial form refers to the physical and spatial characteristics of a city or urban area. This includes the layout, arrangement, and distribution of streets, buildings, open spaces, transportation infrastructure, and other elements that make up the urban environment. Geospatial form plays an important role in shaping the character and functionality of a city, influencing everything from the quality of life of its residents to the economic opportunities available. 
Second,  human mobility refers to the movement of people within and between urban areas. This includes daily activities such as commuting to work or school, as well as less frequent trips for shopping, socializing, and other purposes. Human mobility data can be from city bikes, buses, taxicabs, subways, and mobile devices. Understanding human mobility is essential for urban planners because it affects many aspects of city life, including transportation, land use, and economic development. By analyzing patterns of human mobility, planners can identify where people are going and how they are getting there, and use this information to design more efficient and sustainable transportation systems, locate public services and amenities, and encourage economic activity in areas that are easily accessible.
Third, social interactions refer to the ways in which people interact with each other in urban spaces. This includes both offline social interactions, such as those that take place in  schools and government offices, parks, plazas, and other public spaces, as well as online social interactions, such as those follow, comment, re-twitter activities that take place in Twitter, Snapchat, Facebook. Understanding social interactions is important for urban planners because it affects many aspects of urban life, including social cohesion, community building, and economic development.
Finally, human experts often express planning requirements through natural language or pre-selected options, and the socioeconomic characteristics of an area can have a significant impact on the urban planning process. 
Inspired by the success of deep representation learning, it can be utilized to learn the representations of all geospatial forms, human mobility, social interactions, and human planner instructions. All of these  information representations can be combined to form a conditional generation embedding that can be utilized to generate urban plans that are tailored to the specific needs of the area. 

\noindent\underline{\emph{How to Represent.}} The geographic, mobility, social, human dimensions of a place  we observe in the world is often high-dimensional and complex. We believe there are more meaningful, lower-dimensional representations of that data that capture its underlying structure and patterns. These representations can be learned automatically by machine learning algorithms, rather than being hand-crafted by humans, and can then be used for a downstream land-use configuration generator.
Specifically, we transform these complex and imperfect urban multi-dimensional information into a new representation that captures the most salient features or characteristics of a target area's situational state. 
Representation can be achieved by classic methods, such as, feature extraction, dimensionality reduction, and advanced methods, such as, AutoEncoders, Convolutional Neural Networks, Recurrent Neural Networks, Generative Adversarial Networks. The advanced machine learning based methods, such as deep learning based algorithms, can learn structure knowledge-aware representations that capture and fuse the geospatial,  mobility, social, and human aspects of urban environment into an abstract vector.

\subsubsection{Generation: Learning to Generate  Land-use Configurations}
The representation of all these related information (e.g., geospatial, mobility, social, human textual instruction data)  is to be used as the generation conditions for a deep land-use configuration generative model.
A meaningful land-use configuration generative model includes:  1) neural generative model architecture; 2) objective function of generative model; 3) optimization or training strategy. 

\noindent\underline{\emph{Land-use Configuration Neural Generative Architectures.}} There are several neural architectures that are commonly used in deep generative learning. Below are some of the most popular ones: 1)  Variational Autoencoders (VAEs): VAEs are deep generative models that consist of an encoder network, a decoder network, and a probabilistic latent variable model. The encoder network maps the input data to a probabilistic distribution in the latent space, and the decoder network maps the latent variables back to the input space. The model is trained using variational inference, which involves maximizing a lower bound on the log-likelihood of the data.
2)  Generative Adversarial Networks (GANs): GANs are deep generative models that consist of a generator network and a discriminator network. The generator network generates fake data samples, while the discriminator network tries to distinguish between the fake and real samples. The two networks are trained in a game-theoretic framework, where the generator tries to generate samples that can fool the discriminator, and the discriminator tries to distinguish between the fake and real samples.
3) Autoregressive Models: Autoregressive models are a class of deep generative models that generate data one element at a time, based on the previous elements. The most popular autoregressive model is the PixelCNN, which generates images one pixel at a time, conditioned on the previous pixels.
4) Flow-Based Models: Flow-based models are deep generative models that model the density of the data using a sequence of invertible transformations. The most popular flow-based model is the RealNVP, which models the data using a sequence of affine transformations.
5) Energy-Based Models: Energy-based models are a class of deep generative models that define the probability of a sample based on its energy. The most popular energy-based model is the Boltzmann machine, which consists of a set of binary units that interact with each other based on a learned energy function. With the help of these generative models, we can formulate the urban planning problem into a generative task; that is, given data representations of neighboring environments and human instructions as generation conditions, the model aims to generate an appropriate and optimal land-use configuration option that is tailored to these generation conditions. 

\noindent\underline{\emph{Optimizing Land-use Configuration Generation.}} A generative objective function is essential for optimizing a land-use generative model. This includes both generic  objective  and domain-specific objective. 
Specifically, the objective function of VAE is to maximize the evidence lower bound (ELBO), which is a lower bound on the log-likelihood of the data. The ELBO is defined as the sum of the reconstruction loss (i.e., the negative log-likelihood of the data given the latent variables) and the Kullback-Leibler (KL) divergence between the latent variable distribution and a prior distribution.
In GANs, the objective function is a minimax game between the generator and discriminator networks. The generator network tries to minimize the log-probability of the discriminator network correctly classifying the generated samples as fake, while the discriminator network tries to maximize this log-probability.
In autoregressive models, the objective function is to maximize the log-likelihood of the data given the model parameters. This is typically done by maximizing the cross-entropy loss between the predicted and true values of the next element in the sequence.
In flow-based models, the objective function is to maximize the log-likelihood of the data given the model parameters. This is typically done using the change of variables formula to compute the log-likelihood of the data in terms of the log-likelihood of a simple base distribution.
In energy-based models, the objective function is to minimize the energy of the model given the data. This is typically done using contrastive divergence or Gibbs sampling to estimate the gradient of the energy function with respect to the model parameters.
In the contexts of domain-specific objective, the generative process is subject to domain knowledge, such as, 1) the impacts of the spatial hierarchies between urban functional zones and community configurations; 2) the impacts of the geospatial, mobility, and social connectivity between different areas for urban planning; 3) the diversity and fairness of land-use configurations.

\section{Representative Models of Deep Generative Urban Planning}

Effective urban planning can help mitigate operational and social vulnerabilities in urban systems, such as crimes, traffic congestion, and pollution. Given the high complexity of urban systems, planning tasks are predominantly completed by professional planners, resulting in lengthy planning times. However, recent advancements in deep learning, particularly in deep adversarial learning, offer promising potential for teaching machines to imagine and create. This observation motivates us to reimagine urban planning in the era of AI: What role can deep learning play in urban planning? Can machines develop and learn at human-like capabilities to automatically and swiftly calculate land-use configurations? If so, machines can serve as planning assistants and human planners can adjust machine-generated plans for specific needs.

All of the above evidence indicates that there is a compelling need to develop a data-driven, AI-enabled automated urban planner. However, three unique challenges must be addressed to achieve this goal: 1) How can we quantify a land-use configuration plan? 2) How can we develop a machine learning framework that can learn from the good and bad of existing urban communities in terms of land-use configuration? 3) How can we evaluate the quality of generated land-use configurations?

\subsection{Automated Urban Planner via Generative Adversarial Learning}
\subsubsection{Research Gap}

In the following sections, we will present our research insights and proposed solutions to address these challenges.

\subsubsection{Key Perspectives}
\begin{figure}[!thbp]
    \centering
    \includegraphics[width=0.8\linewidth]{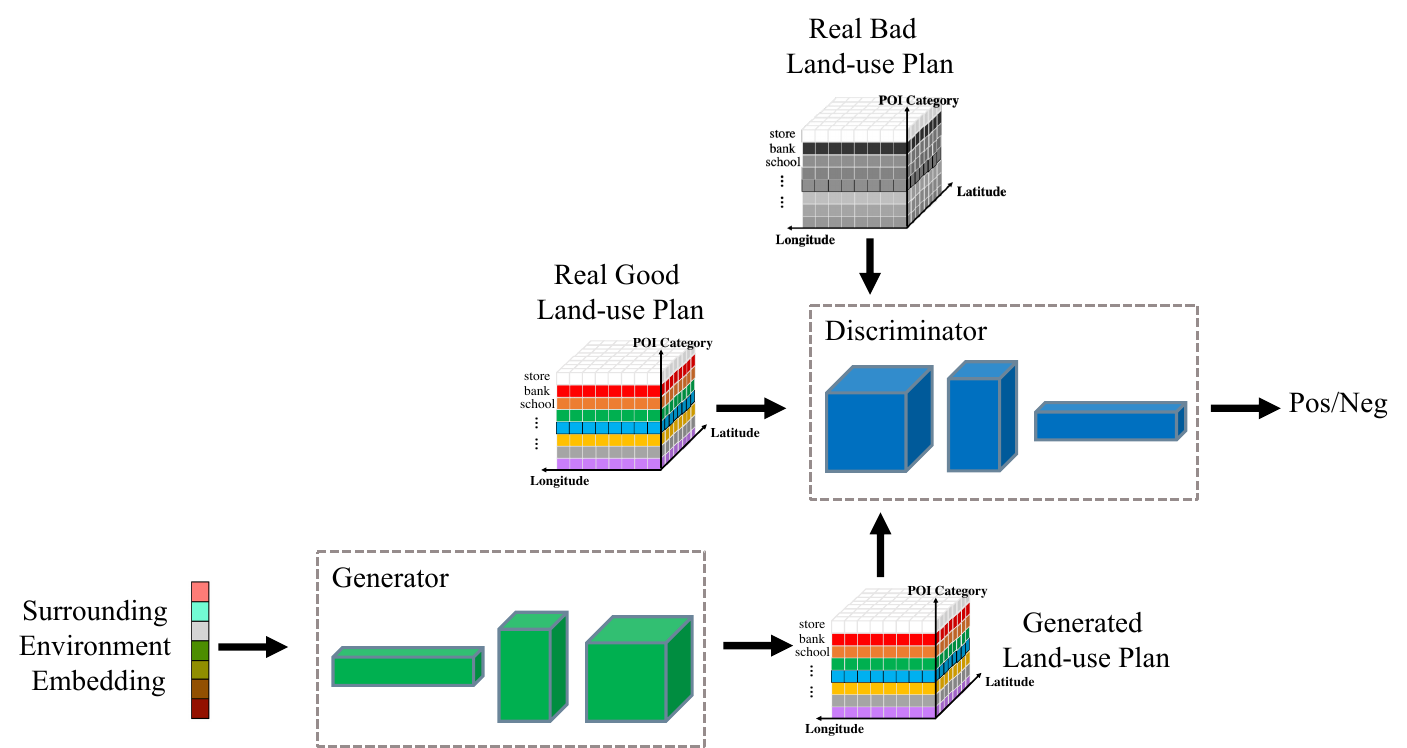}
    \caption{An overview of the LUCGAN framework: This model comprises two primary components. Firstly, the generator endeavors to create optimal land-use configurations by utilizing the embedding of the surrounding environment. Secondly, the discriminator's objective is to assign elevated scores to real favorable configurations, while attributing lower scores to both unfavorable real and generated configurations.}
    \label{lucgan}
     \vspace{1.7cm}
\end{figure}

We propose an automated urban planner, namely LUCGAN, to resolve these challenges in order to generate effective land-use configurations based on the planning constraints of surrounding environments~\cite{gan_planner,gan_planner_tsas}, as shown in \textbf{Figure~\ref{lucgan}}.
Our key insights can be summarized as follows:
First, we develop a novel approach by quantifying a land-use configuration plan using a latitude-longitude-channel tensor. This representation enables us to model the complex spatial relationships that exist in urban environments.
Second, we propose a socioeconomic interaction perspective to better understand the optimization of the coupling between a community and its surrounding environments in urban planning processes. This perspective allows us to better integrate social and economic considerations in the development of land-use configurations.
Third, we rethink the automated urban planning problem as an adversarial learning framework, in which a machine generator maps surrounding spatial contexts into a configuration tensor. This framework enables us to learn from large-scale, heterogeneous datasets and generate high-quality land-use configurations.
Finally, we evaluate the effectiveness of our proposed approach by conducting extensive experiments and visualizations with real-world data. Although evaluation is challenging, we demonstrate the value of our method through multiple aspects, such as quantitative metrics and qualitative analysis.

\subsubsection{Summary of Proposed Solution}
We propose an automated urban planner framework, namely LUCGAN, consisting of two phases: 1) surrounding contexts embedding and 2) land-use configuration generation. In the embedding phase, we extract explicit features of the surrounding contexts, such as value-added space and POI distribution, and construct a spatial attributed graph by mapping the features to eight vertices that model the eight geographical squares of the surrounding contexts. A graph embedding model is then applied to preserve the graph information into an embedding vector, which represents the entire information of the surrounding contexts. In the land-use configuration generation phase, the embedding vector of the contexts is inputted into an extended generative adversarial network (GAN) to learn and gradually formulate the distribution of well-planned land-use configurations instead of poorly-planned configurations. When the model converges, the extended GAN generates suitable and desired land-use configurations based on the embeddings of the surrounding contexts.

\subsubsection{Limitations}
While LUCGAN has demonstrated the ability to generate land-use configurations based on contextual information, it falls short in its capacity to incorporate the planning requirements set forth by human experts. Due to the lack of control over the generation process, the model struggles to grasp the semantic planning meanings and human-centric considerations necessary for producing land-use configurations that align with expert guidelines.

\begin{figure}[t]
    \centering
    
    \includegraphics[width=0.8\linewidth]{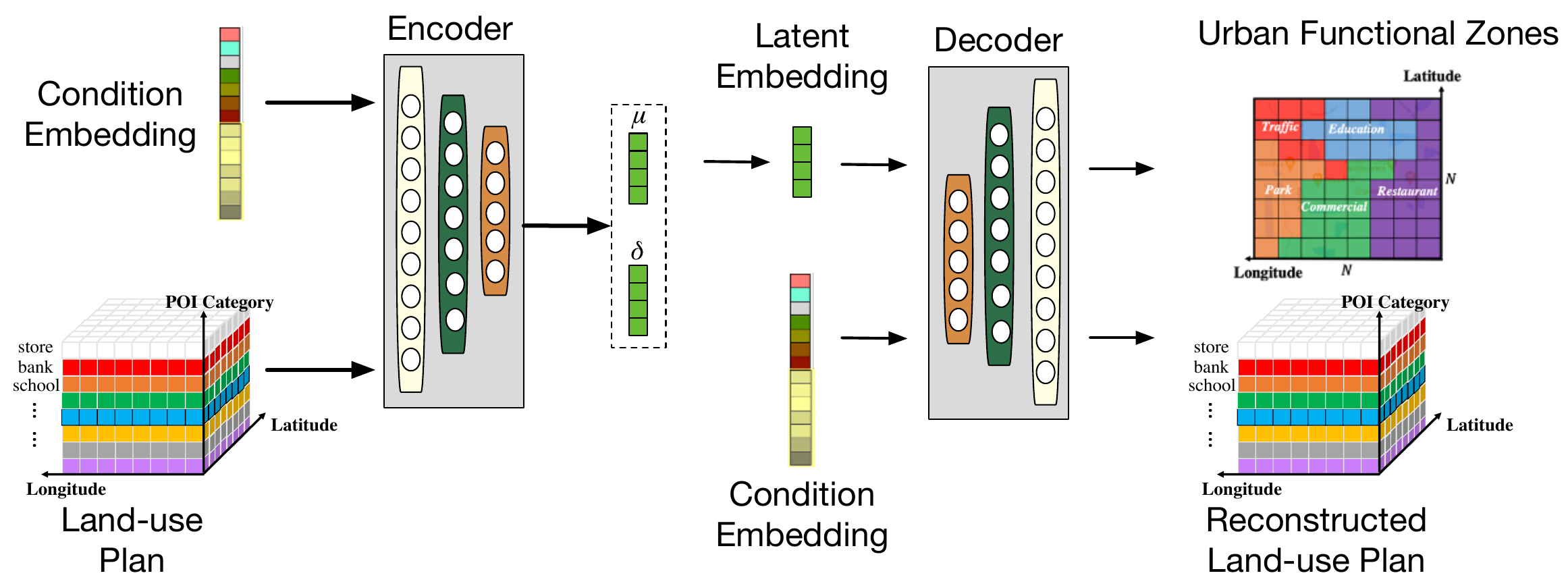}
    \vspace{0.1cm}
    \caption{An Overview of the CLUVAE Framework: This architecture encompasses two primary components. Firstly, the encoder is designed to learn the distribution of information pertaining to conditional embeddings and land-use plans. Secondly, the decoder is tasked with reconstructing urban functional zones and land-use plans, thereby facilitating the generation of accurate and optimal outcomes.}
    \label{cluvae}
\end{figure}

\subsection{Automated Urban Planner via Variational Autoencoder}

\subsubsection{Research Gap}
Previous research in the field has generally overlooked the importance of personalized human guidance in the planning process, as well as the need for a hierarchical spatial structure to guide planning generation. Additionally, an absence of large quantities of land-use configuration samples presents a significant data sparsity challenge. To address these issues, we propose a novel deep human-guided urban planning method that offers a comprehensive solution to the aforementioned challenges~\cite{vae_planner,wang2023automated}.

\subsubsection{Key Perspectives}
We propose a deep conditional variational encoder-decoder framework, namely CLUVAE, to formulate automated urban planning, which accepts human textual information and the surrounding environment as input and outputs the desirable land-use configurations, as shown in \textbf{Figure~\ref{cluvae}}.  To overcome the sparsity issue of land-use configuration data, the encoder returns a distribution over the learned latent embedding space, which improves the model's robustness and diversity. Additionally, the decoder reconstructs the urban function labels and POI distribution tensor, capturing the spatial hierarchies between different zones and communities. These features give human experts greater control over the generation process and produce high-quality land use configurations that reflect their needs and preferences.

\subsubsection{Summary of Proposed Solution}
The proposed CLUVAE framework includes four key steps that enable effective automated urban planning. Firstly, a dataset of land-use configurations and corresponding urban functional zones in various target areas is collected. Secondly, embeddings of surrounding environment characteristics and human guidance are obtained. To formulate the former, spatial attributed graphs are used, and a graph autoencoder model is employed to obtain its embedding vector. The latter is represented as a one-hot vector of the human guidance text. These embeddings are concatenated together to form the conditional embedding of the generation model. The third step involves building a deep robust land-use configuration generation model. The encoder of the generation model learns the distribution that reflects the correlation between the conditional embedding and the corresponding land-use configuration. The decoder of the generation model reconstructs land-use configurations and urban functional zones based on the condition embedding and the latent embedding sampled from the previously learned distribution in order to capture spatial hierarchies. Finally, the well-trained decoder serves as the desired land-use configuration generator.

\vspace{0.6cm}
\subsubsection{Limitations}
The CLUVAE framework represents a promising approach for generating desirable land-use configurations by incorporating human expertise and surrounding contexts while capturing spatial hierarchies in urban planning through a regularization item. However, the generation performance of the model is currently unstable due to two underlying reasons. Firstly, the regularization item may not comprehensively capture the spatial hierarchies, and its contribution is difficult to adjust during the optimization process. Secondly, the model may fail to capture the impacts of the urban plans between different geographical areas. These limitations require further research and development to improve the performance and robustness.

\subsection{Human-Environment and Spatial-Hierarchy Aware  Automated Urban Planner via Transformer}
\subsubsection{Research Gap}
Existing research has yielded encouraging outcomes; however, these studies have overlooked the dependencies and reciprocal impacts among various subarea planning efforts, failing to fully encapsulate the hierarchical relationships between coarse-grained urban functional zones and fine-grained land-use configurations. Moreover, accurately interpreting human instruction remains a significant hurdle. Consequently, in order to confront the shortcomings inherent in current automated urban planning literature, we propose a transformer-based automated urban planner that is capable of comprehending human-environment constraints and maintaining cognizance of spatial hierarchies within the urban planning process~\cite {transformer_planner}.

\begin{figure}[t]
    \centering
    \includegraphics[width=1.0\linewidth]{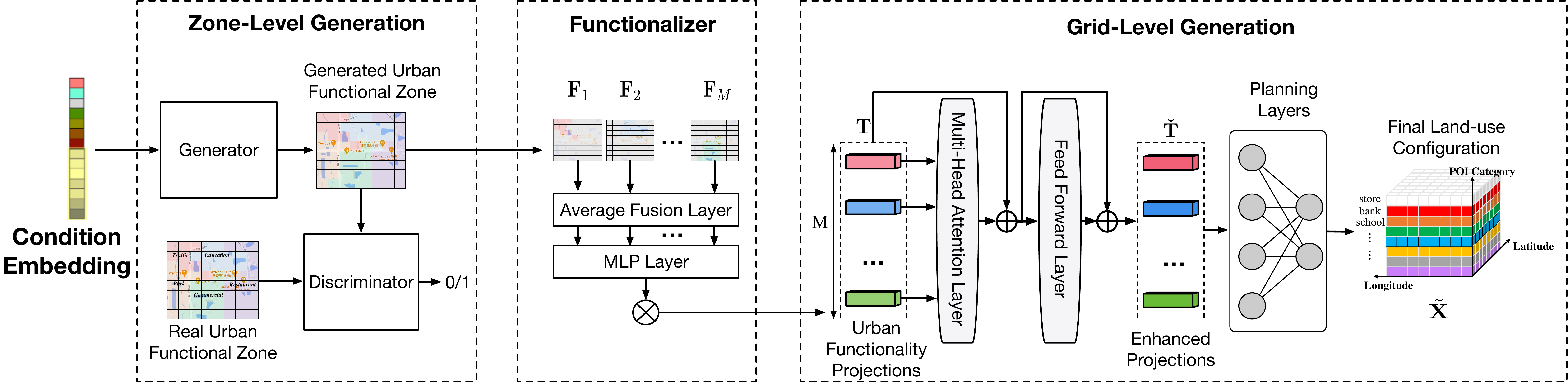}
    \vspace{0.1cm}
    \caption{An Overview of IHPlanner: This framework is inspired by the workflow of human urban experts. Firstly, the zone-level generation module creates a coarse outline of an urban plan. Secondly, the grid-level generation component refines the land-use configurations. During the process, the Functionalizer is introduced to comprehensively understand the planning requirements of human experts and surrounding environments.}
    \label{ihplanner}
    \vspace{0.6cm}
\end{figure}

\subsubsection{Key Perspectives}
We present an innovative, human-guided, deep hierarchical generative approach to automated urban planning, incorporating spatial hierarchical associations and planning dependencies to address existing limitations in the literature. Our proposed framework, namely IHPlanner, emulates the traditional urban planning workflow through two generative stages: an initial rough urban plan sketch, followed by a refinement process yielding fine-grained results, as shown in \textbf{Figure~\ref{ihplanner}}. This formulation enables the automated urban planner to fully capture spatial hierarchies within urban planning.
Furthermore, we introduce a Functionalizer module designed to thoroughly understand and integrate human instructions and surrounding environmental factors, ensuring that IHPlanner comprehends the planning requirements necessary to generate optimal land-use configurations. Additionally, a multi-attention mechanism is employed to capture the influences and impacts among different subareas, facilitating the generation of more plausible land-use configurations.

\subsubsection{Summary of Proposed Solution}
Our pipeline framework comprises four primary components: conditioning augmentation, zone-level generation, Functionalizer, and grid-level generation. Initially, for an empty target area, we embed human instructions and surrounding environmental factors into a vector representation to preserve planning requirements. To address data sparsity, the conditioning augmentation module enhances data diversity. Subsequently, the zone-level generation module produces a zone-level plan, laying the planning foundation for the grid-level generation. Within the Functionalizer module, the semantics of planning requirements are projected onto different functional zones, resulting in urban functionality projections and converting planning dependencies across functional zones into semantic correlations. Lastly, the grid-level generation module employs multi-attentions to capture these semantic correlations and utilizes planning layers to generate grid-level planning.

\subsubsection{Limitations}
Despite the promising performance of IHPlanner in automated urban planning, certain limitations warrant consideration. First, real-world planning requirements posed by human experts are considerably more complex than what IHPlanner currently addresses, which relies on extracting key information from natural language and representing it with one-hot vectors. Second, the proposed framework has not been extensively tested in real-world scenarios due to associated costs, necessitating the development of a simulation platform to bridge the gap between automated urban planners and practical applications. Third, the evaluation metrics employed for automated urban planning are overly simplistic, primarily focusing on distribution similarity. To enhance objectivity, it is advisable to incorporate multiple professional urban planning metrics in the evaluation process.

\subsection{Suitability of the Models Across Different Urban Planning Contexts}
The effectiveness of deep generative models in urban planning depends on the specific task they are used for. GANs are great at automating zoning and land-use planning, as they can generate realistic layouts by learning from existing urban data. However, they need large datasets and careful validation, making them less practical in cities with changing regulations. VAEs, on the other hand, are better suited for scenario-based planning because they can generate multiple land-use options and handle incomplete or sparse data. This makes them useful for testing different urban development possibilities and planning for uncertainty. Transformer-based models shine in human-in-the-loop planning and policy-driven urban design. These models can process complex spatial relationships and incorporate real-time feedback from urban planners, making them highly interactive and adaptable. While each model has its strengths, none are perfect on their own. Their success depends on how well they integrate planning expertise, ensure transparency, and align with broader sustainability goals. Moving forward, a hybrid approach—combining the predictive power of GANs, the flexibility of VAEs, and the adaptability of transformers—could help bridge the gap between AI-driven insights and real-world urban planning needs.

\section{Envision The Future of Urban Planning: Automated, Generative, Geospatial, Social, Economic, Environmental Knowledge-guided, Human-machine Collaborative, and Fairness-aware}

\begin{figure}[t]
    \centering
    \includegraphics[width=0.85\linewidth]{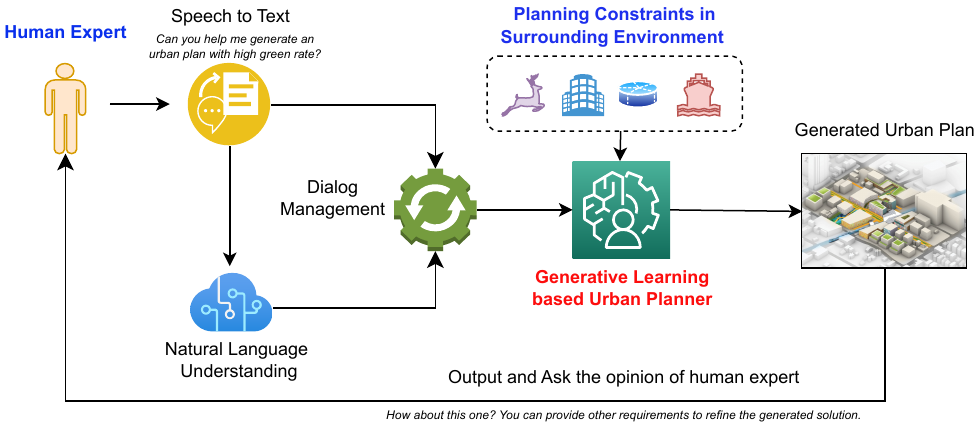}
    \caption{A future-oriented urban planning framework in which human experts engage with a speech-driven, generative learning-based urban planner. During each iteration, experts convey their planning requirements, and upon understanding the semantic implications, the planner generates corresponding urban plans, taking into account existing planning constraints in the surrounding environment. Subsequently, the devised plans are presented to the experts for soliciting refinement feedback. The iterative process continues with experts providing additional planning requirements until the urban plan achieves comprehensive satisfaction.}
    \label{future_plan}
\end{figure}

With the increasing availability of data, urban planners can use data analytics to better understand how people use urban spaces, identify patterns and trends, and make informed decisions about future urban planning. 
Besides, mixed-use development involves creating spaces that combine residential, commercial, and recreational uses. This can help to create more livable, walkable communities that reduce the need for car travel.
Based on that, future urban planning is likely to be influenced by the rapid advancements of  data-centric AI techniques and model-centric AI techniques. 
The data-centric AI techniques include spatial-temporal representation~\cite{wang2021reinforced,wang2018learning,wang2019spatiotemporal} and  multi-modality learning~\cite{korot2021code,chen2022simple,lu2022learning}. 
The model-centric AI technologies include deep generative learning~\cite{goodfellow2020generative,ho2020denoising}, pre-training~\cite{kenton2019bert}, conversational AI~\cite{bubeck2023sparks,brown2020language}, and reinforcement learning with human-in-the-loop feedback~\cite{arakawa2018dqn}. 
As a result, \emph{we envision that future urban planning will be automated, generative, geospatial, social, economic, environmental knowledge-guided, human-machine collaborative, and fairness-aware}, as shown in \textbf{Figure}~\ref{future_plan}.
These features can help to ensure that the designs are effective and meet the needs of the community.

\begin{itemize}
    \item \emph{Automated:} "Automated" typically refers to the use of AI to streamline or augment the planning process. This may include the use of computer programs, algorithms, or other digital tools to help analyze data, generate design options, or make decisions about urban development. Automated urban planning can potentially help planners to more efficiently and accurately evaluate different scenarios and outcomes, and to incorporate more complex and diverse factors into their decision-making processes. However, it is important to ensure that such tools are designed and implemented in a way that is transparent, equitable, and respectful of community input and engagement.

    \item  \emph{Generative:}  "Generative" refers to the use of generative models in deep learning to simulate and create new urban design possibilities. These models are trained on large datasets of urban features and patterns, and can be used to generate new designs or simulate the outcomes of different design choices. However, generative models may have limitations in terms of their ability to capture the full complexity of urban systems and the social and cultural factors that shape them.

    \item \emph{Human-machine Collaborative: } "Human-machine collaborative" refers to an approach to urban planning that involves learning to generate urban planning with human feedback in the loop. This is a ChatGPT-like conversational collaboration   between humans and AI. This approach seeks to combine the creativity, expertise, experiences, intuitions of human designers into intelligent machines. This can potentially lead to more efficient, effective, and innovative urban designs that better meet the practical needs of diverse communities.

    \item \emph{Geospatial, Mobility, Social, Economic, Environmental Knowledge-Guided: } 
Urban systems involves five important dimensions: geospatial, mobility, social, economic, and environmental dimensions. Domain knowledge regularization or guidance are critical to produce meaningful and useful generative planning because: 1) deep generative planning is often trained on large urban data that are noisy or incomplete, meaning that the patterns and relationships they learn may not accurately reflect the underlying structure of land-use configuration and relationships to other dimensions. This can lead to the generation of unrealistic or meaningless outputs. 2) deep generative planning may not have a full understanding of the entire context (geospatial, social, economic, mobility, and environmental) in which traditional urban planning will be used. Incorporating domain knowledge or guidance can help ensure that the generated outputs are both realistic and useful. 3)  deep generative plan can be prone to producing land-use configurations that reinforce existing biases or inequities in existing community land-use configurations. Incorporating domain knowledge or guidance can help mitigate this risk.

    \item \emph{Human-centric Planning with Fairness Awareness: } Fairness refers to considerations of equitable access to affordable housing, transportation, public services, and green spaces, as well as other resources (e.g., healthcare resources) over different sub populations (e.g., aging groups). As AI becomes increasingly integrated into urban planning, ensuring responsible AI practices is essential for promoting fairness, mitigating biases, enhancing transparency, and safeguarding community interests~\cite{ye2025artificial}. In this regard, responsible AI frameworks can help achieve ethical considerations, explainability, and accountability, ensuring that AI-driven urban planning tools align with societal values and do not disproportionately disadvantage marginalized communities~\cite{yigitcanlar2024unlocking}. This also calls for strong governance and clear regulations to guide the collaboration between AI-driven planning systems and human experts~\cite{brand2023ensuring}.

\end{itemize}

\subsection{Future Human-Machine Collaborative Planning: Integrating Generative Intelligence and  Human Feedback in The Loop}

\noindent\underline{\emph{Why Integrating Generative Intelligence and Human Feedback in The Loop?}} Although human-level AI does not exist  to replace urban planning experts, we envision a bold perspective: human-machine collaborative planning, regarding future urban planning. This new human-machine collaborative planning platform will integrate both deep generative intelligence and human feedback in the loop. The underlying benefits of such integration are: 
\emph{1) automating land-use planning}: Deep generative planning can be used to generate land-use configuration alternatives based on planning conditions, such as site characteristics, building codes, neighborhood environments, and sustainability goals, by learning from historical land-use configuration data. 
\emph{2) refining design options via human feedback in the loop fine tuning:}
Once generated, land use configurations can be presented to human planners to collect human feedback to refine design alternatives. If a planner identifies a particular design aspect (e.g., working distances, green rates)  that needs improvement, the feedback can be used to incentivize deep generative planner to generate new alternatives that address that concern.
\emph{3) mutual enhancement in creativity:}
During such interactive, conversational, and iterative design process, machine planners can present new ideas and concepts that human planners may not have considered otherwise. This can help human planners to provide more quality human feedback outside the box.
\emph{4) balancing multiple perspectives:} Deep generative planning can be used to balance multiple perspectives and considerations into the design process. 

\noindent\underline{\emph{A Deep Generative Planning with Human Feedback in The Loop Perspective.}}
Inspired by ChatGPT~\footnote{\url{https://openai.com/research/instruction-following}}, \textbf{Figure \ref{future_plan}} envisions a  human-machine collaborative planning framework to integrate both deep generative urban planning and human feedback in the loop as one. 

\begin{itemize}
    \item \emph{Training Deep Generative Planner.} 
    \begin{itemize}
        \item Pre-training: In this stage, we train a deep generative planning model (decoder-only transformer) on the  land-use configuration data of communities and related urban geography, human mobility, and location based social media data. The backbone neural network architecture of deep generative planning can be auto-encoder, GAN, generative flows, decoder-only transformer. 
        The objective is to train a generative model that can predict a land-use configuration plan given a place and corresponding  geography, mobility, environment, economic, and social media information in a way that is geographically and functionally meaningful to fit into  the place. After the pre-training stage, the model can generate a land-use configuration given a place and the place's related information, but it is not capable of responding to questions. 
        \item Fine-tuning: This stage is a three-step process that turns the pre-trained land-use generative model into a conversational model with human feedback in the loop abilities: 
        \begin{itemize}
            \item  Step 1: Collect training data (the pairs of planning improvement requests and re-generated land-use configurations), and fine-tune the pre-trained deep generative planning model on this data. The deep generative planning takes a planning improvement request as new input generation conditions and learns to generate a land-use configuration similar to the training data.  
            \item Step 2: Collect more data (the pairs of planning improvement requests and regenerated land-use configurations) and train a reward model to rank these answers from most relevant to least relevant.
            \item Step 3: Use reinforcement learning (e.g., PPO optimization) to fine-tune the model so the deep generative planning model's generated land-use configurations are more accurate. 
        \end{itemize}
    \end{itemize}

    \item \emph{Answering A Planning Improvement Request.} 
        \begin{itemize}
        \item Step 1: The user engages with the generative learning-based urban planning system by articulating planning prerequisites via spoken language. For example, "Could you assist in generating an urban design that incorporates a substantial degree of green spaces?"
        \item Step 2: The posed inquiry is initially converted into textual form and subsequently processed through a natural language understanding model to discern its semantic content. Simultaneously, the information is incorporated into a dialogue management module to maintain a record of the present conversational state.
        \item Step 3: The generative learning-based urban planning system assimilates information from the current dialogue and the contextual planning constraints of the surrounding environment. It subsequently generates an urban plan that fulfills the specified planning requirements.
        \item Step 4: The produced urban plan is presented to the human expert for the purpose of soliciting their evaluation and identifying potential areas for refinement or enhancement.
        \item Step 5: Human experts may offer specific recommendations for improvement, enabling the iterative refinement of the generated outcome until the optimal solution is ultimately achieved.
        \end{itemize}
\end{itemize}

In order to actualize a human-machine collaborative planning framework, we must confront various technical hurdles. These challenges present a multitude of promising research avenues that hold substantial potential for advancing the field in the future. Moreover, we also note that beyond conventional AI-driven automation, there are recent considerations to engage 'Agentic AI' that has the potential to transform urban planning by enabling systems that actively make decisions, adapt through learning, and refine urban designs in response to changing conditions~\cite{de2025responsible}. Unlike passive generative models, Agentic AI operates with goal-oriented flexibility, allowing it to assess urban scenarios, suggest planning interventions, and continuously improve solutions through iterative collaboration with human experts--human in the loop~\cite{qian2023ai}. 

\subsubsection{Trend 1: Automation of High-Quality Data Preparation}
Deep generative AI has demonstrated remarkable human-like intelligence in generating intricate solutions contingent upon the input conditions, as evidenced by image generation~\cite{oussidi2018deep}. Constructing a superior deep generative system necessitates the acquisition of large quantities of high-quality and effective data, which poses a significant challenge. In the context of urban plan generation, it is essential to delve deeper into the following three directions of investigation:

\noindent\textbf{Integrating Geospatial Hierarchy in Urban Planning.} 
In the domain of urban planning, community-level planning (land-use configuration) is impacted by zone-level planning (urban functional zones). Drawing a parallel with artistic illustration, urban functional zones can be likened to an initial sketch, whereas land-use configurations represent comprehensive planning solutions. Recognizing these dependencies can alleviate the generative complexities inherent in deep generative models, ultimately leading to more efficient and ideal land-use configurations.
We have attempted to incorporate these hierarchical relationships through two approaches:
1) Integrating structured regularization into the land-use configuration generation process~\cite{vae_planner};
2) Developing a two-stage generation framework that imitates the workflow of human experts~\cite{transformer_planner,hu2024dual}.
However, it raises the question: Is there a more effective method to assimilate such hierarchical structures into the planning generation process? A potential avenue to explore involves devising a more intelligent approach to learning and incorporating this structural information, ultimately enabling the generative model to produce more plausible outcomes.

\noindent\textbf{Overcoming Data Sparsity in Urban Planning.} 
Urban planning data are often costly, imperfect, and sparse. Even in a large city such as Beijing, China, there are merely around 3,000 residential communities, resulting in a scarcity of high-quality training data. Consequently, addressing the following question becomes essential: How can we make a generative planning model with resilience against data sparsity?
One approach to mitigate the sparsity issue in land-use configuration data is to force the encoder to return distribution over the latent embedding space, rather than a singular embedding point. Such a distribution can more accurately approximate the data dispersion pattern within the embedding space, thereby allowing for continuous sampling from the distribution without necessitating extensive, high-quality training land-use configuration data. Moreover, this strategy introduces randomness in embedding, diversifies representations, and enhances model robustness.
However, are there more effective solutions to address the data sparsity problem? Presently, we assume that the distribution of urban data follows a normal distribution, yet in reality, this distribution is considerably more complex. As such, innovative solutions are warranted to augment data volume and diversity, enabling the deep generative model to better comprehend the characteristics of land-use configurations and ultimately yield superior outcomes.

\noindent\textbf{Automated Training Data Preparation and Pretraining for Deep Generative Planning.}  
Pre-training constitutes the process of training an extensive model on an abundant dataset, aiming to capture general patterns and representations that can be repurposed across a variety of tasks. 
In the context of deep generative learning, pre-training involves employing unsupervised learning techniques to initialize the weights of deep neural networks. The objective of pre-training is to ascertain a suitable set of initial weights that expedite convergence and improve performance on specific generative tasks. A prevalent pre-training approach in generative learning involves the utilization of autoencoding, an unsupervised learning technique. In an autoencoder, the network is trained to reconstruct its input data by learning a compressed representation within a lower-dimensional latent space. The encoder maps the input data to the latent space, while the decoder performs the inverse operation. By minimizing the discrepancy between the input and reconstructed data, the network captures critical data features within the latent space. Once trained, the encoder weights can initialize deep generative models, such as generative adversarial networks (GAN)\cite{goodfellow2020generative} or variational autoencoders (VAE)\cite{kingma2019introduction}. 
Commencing with pre-learned weights capturing vital features allows the generative model to converge rapidly and achieve superior performance.
The diversity, scale, and comprehensiveness of the training data are crucial for pre-training deep generative learning models. We envision reinforcement learning as an instrument for automated training data preparation. In essence, we anticipate the development of reinforcement agents capable of emulating the generative process of placing various buildings within a region. As reinforcement agents automate community configuration exploration, they generate diverse data. Although conceptualizing urban planning as a generative task facilitates optimization for effective generative model parameter learning and novel configuration planning reconstruction, an alternative perspective involves formulating urban planning as a reinforcement learning actor-critic structure. This approach continuously explores various urban community configurations, generating future pseudo-training data for pre-training deep generative planning models.

\subsubsection{Trend 2: Informative Urban Community Configuration Representation}
Urban community configuration pertains to the systematic arrangement and organization of structures, thoroughfares, and public spaces within metropolitan areas. This process encompasses the planning and design of the constructed environment, including the positioning of buildings, open areas, and infrastructural elements.
The objective of urban community configuration is to establish a functional, aesthetically pleasing, and sustainable urban setting that accommodates the requirements of inhabitants and enterprises. This may entail devising mixed-use developments incorporating residential, commercial, and recreational purposes, crafting pedestrian-friendly districts that encourage physical exercise and minimize automobile dependency, and generating public spaces that stimulate social interaction and community involvement.
Consequently, the inquiry arises: how can we employ a more machine-comprehensible representation to characterize the relational structure of intricate community configurations, encompassing edifices, their functions, quantities, distances, and additional elements?
Numerous extant studies utilize a high-dimensional tensor to represent locations, building functionality classifications, building counts, and other pertinent attributes. Nonetheless, how do these dimensions engage with each other spatially, socially, functionally, and semantically within the context of community configurations?

\noindent\textbf{Community Configuration as Attributed Graphs.} 
We envision that future representations of urban community configurations can be approached through a novel perspective: community configuration as an attributed graph.
An attributed graph is a structure wherein each vertex or edge is associated with a set of attributes or properties. These attributes are characteristics or features describing geospatial entities within the graph, providing supplementary information. In an attributed graph, vertices can possess attributes such as names, labels, types, locations, or other properties, while edges may have attributes like weights, lengths, directions, or relationship types. Depending on the granularity of community configuration, these attributes can be represented as key-value pairs or complex data structures.
In the context of urban community configurations, an attributed graph can depict a spatial region where vertices represent buildings and edges signify relationships among them. Attributes can describe building aspects such as size, age, height, design, materials, energy efficiency, accessibility, safety, amenities, functional categories, or their influence on other structures within the network. Subsequently, these attributes can be employed to train deep generative neural networks for generating new urban community configurations. Attributed graphs offer a flexible and expressive framework for representing intricate urban community configuration structures with rich properties, resulting in more practical, attributed graph-based community configurations.

\subsubsection{Trend 3: Fairness-aware Urban Planning}

In fact, the objective of urban planning is to harmonize the relationship between human experts and the environment.
Achieving various forms of fairness in urban planning is crucial to ensure the development of the urban plan and balance the relationship between humans and the environment~\cite{yigitcanlar2024unlocking}. The following potential directions warrant further exploration:

\noindent\textbf{Functionality fairness oriented planning.}
Functionality fairness-oriented planning is a vital aspect of urban planning that must be considered in the future~\cite{brand2023ensuring}. One of the challenges in urban planning is ensuring that every city area has equal access to essential services such as healthcare, education, and public transportation~\cite{mehrabi2021survey,zhou2021transportation,ahmad2020fairness}. Automated urban planning can employ sophisticated algorithms to allocate resources more efficiently and guarantee that each city area receives adequate attention. Furthermore, these algorithms can be used by planners to establish a more equitable distribution of job opportunities and housing options, preventing poverty concentration in specific areas.
To accomplish functionality fairness in automated urban planning, it is critical to analyze demographic data using machine learning algorithms, identifying city areas with higher needs for essential services like healthcare, education, and public transportation. Moreover, simulation tools can be created to predict the effects of new developments on various city areas and allocate resources accordingly, ensuring an equitable distribution of new development. Additionally, machine learning algorithms can monitor job opportunities and housing option distributions, adjusting planning algorithms to avoid poverty concentration in particular areas. These measures guarantee that every city area receives sufficient attention and resources, promoting equitable distribution of job opportunities and housing options, and preventing poverty concentration in specific areas. By prioritizing functionality fairness, automated urban planning can create more livable and equitable cities that cater to the needs of all residents.

\noindent\textbf{Mobility and accessibility fairness oriented planning.}
Promoting mobility and accessibility fairness is a vital aspect of future urban planning endeavors. With cars maintaining a dominant role in transportation, city planners need to advocate for alternative means such as bicycles, electric scooters, and public transportation~\cite{wang2021measuring}. Automated urban planning can leverage cutting-edge algorithms to optimize the placement of transportation infrastructure, guaranteeing access to public transportation across the city. Furthermore, it can support the development of pedestrian-friendly streets and bike-friendly infrastructure, consequently lowering greenhouse gas emissions, bettering air quality, and stimulating healthier lifestyles.
To achieve mobility and accessibility fairness in automated urban planning, it is essential to create transportation demand models by employing machine learning algorithms. These algorithms analyze mobility patterns and detect areas lacking adequate public transportation access. Additionally, simulation tools can be utilized to examine the effects of new transportation infrastructure and establish the most efficient infrastructure locations, ensuring equitable access for all residents. Algorithms can be refined to strategically position bike lanes and pedestrian crossings, enhancing the city's bike and pedestrian-friendliness and advocating for sustainable transportation. Implementing these measures guarantees widespread public transportation access and nurtures alternative transportation methods, ultimately alleviating traffic congestion and improving air quality. By emphasizing mobility and accessibility fairness, automated urban planning can foster more sustainable and livable cities that address the transportation needs of all residents.

\noindent\textbf{Environment and green-oriented planning.}
Environmental and green-oriented planning is a critical component of future urban planning. As urban growth continues, the importance of environmentally sustainable practices, such as energy-efficient buildings, green spaces, and renewable energy sources, becomes increasingly apparent. Automated urban planning can employ advanced algorithms to optimize green space placement and encourage energy-efficient building development. Furthermore, it can facilitate the implementation of renewable energy sources like solar and wind power, reducing greenhouse gas emissions and fostering sustainable living. Ultimately, the future of automated urban planning must consider environmental sustainability, mobility and accessibility, and functionality fairness, ensuring equitable, sustainable, and livable cities for all residents.
To achieve environmental sustainability in automated urban planning, it is crucial to analyze energy usage patterns with machine learning algorithms, identifying areas of the city with high energy consumption and prioritizing energy-efficient buildings within those regions. Additionally, simulation tools can be used to forecast the environmental impact of new buildings and developments, adjusting planning algorithms to emphasize sustainability. Moreover, algorithms can be designed to track green space distribution and optimize park and green space placement, ensuring city-wide access. These measures guarantee environmentally sustainable urban development, promote energy efficiency, reduce greenhouse gas emissions, and enhance green space availability, contributing to healthier and more livable cities. By focusing on environmental sustainability, automated urban planning can establish a more resilient and sustainable urban environment that accommodates the needs of both present and future generations.

\section{Conclusion and Research Directions}
Urban planning is essential for addressing the challenges of growing urban populations, promoting sustainable development, and enhancing the quality of life in cities. This paper has examined the complexities of urban planning and the potential of AI-assisted tools to transform the field, making it more efficient and equitable.
Four key trends will drive the future of AI-assisted urban planning: new configuration representation, new generative learning, human-machine collaborative planning via conversational AI, and fairness-aware planning. By embracing these trends, we can ensure that our urban planning methodologies are more adaptable, innovative, and inclusive.
First, new configuration representation techniques will enable urban planners to model and simulate urban environments with greater accuracy and context awareness. These methods will facilitate better decision-making by incorporating diverse perspectives and generating solutions tailored to specific urban settings.
Second, new generative learning approaches will empower urban planners to develop more sophisticated and creative designs by leveraging AI's ability to analyze vast amounts of data and identify patterns that may elude human understanding. This will not only enhance the efficacy of urban planning but will also foster innovation and sustainability.
Third, human-machine collaborative planning via conversational AI will create a synergistic relationship between urban planners and AI-assisted tools. This collaboration will facilitate more effective problem-solving by combining the strengths of human expertise with AI's analytical capabilities, ultimately leading to more informed and balanced decision-making.
Finally, fairness-aware planning will ensure that AI-assisted urban planning takes into account the needs and aspirations of all citizens, reducing the risk of biased or discriminatory outcomes. By integrating fairness and equity into the core of AI-assisted urban planning, we can work towards building cities that are more inclusive and just.
As urban populations continue to grow and urban planning challenges become increasingly complex, AI-assisted urban planning offers an innovative solution to create sustainable, resilient, and equitable urban spaces. By adopting the trends outlined in this paper and fostering a multidisciplinary approach combining computer science and urban planning strengths, we can rethink our cities' future and contribute to a more prosperous, inclusive, and sustainable world.
Building on the insights generated in this study, prospective research should focus on developing robust frameworks that support real-world deployment of automated urban planners. This includes integrating Agentic AI to enable adaptive, goal-driven planning systems and enhancing conversational AI capabilities for more intuitive human-AI collaboration. A pressing challenge lies in overcoming data limitations through simulation-based data generation and transfer learning. Moreover, embedding equity and transparency into generative planning models will be crucial for ensuring socially responsible outcomes. Novel evaluation metrics that align with professional planning standards must be developed to better assess the quality and fairness of AI-generated plans. Cross-disciplinary collaborations involving urban planners, AI researchers, and policymakers are essential to co-design frameworks that are both technically sound and ethically grounded. Furthermore, advancing from theory and model development to application will require robust simulations, field validations, and regulatory innovation to embed these technologies into future planning practices. 
We are going through disruptive times marked by rapid and exponential technological development, affecting all sectors—including urban planning. Advancing research in this field is not only about improving planning mechanisms but also about delivering timely solutions to the colossal challenges facing our cities—before it is too late.
\vspace{0.1cm}


\bibliographystyle{ACM-Reference-Format}
\bibliography{snbibliography}

\end{document}